\documentclass{article}

\usepackage{microtype}
\usepackage{graphicx}
\usepackage{subfigure}
\usepackage{booktabs} 

\usepackage{hyperref}

\usepackage[accepted]{icml2019}

\icmltitlerunning{}

\begin{document}

\twocolumn[
\vspace{-20pt}
\icmltitle{\vspace{-6pt}A Bi-Directional Co-Design Approach to Enable Deep Learning \\ on IoT Devices\vspace{-6pt}}
\vspace{-8pt}


\icmlsetsymbol{equal}{*}

\begin{icmlauthorlist}
\icmlauthor{Xiaofan Zhang}{ui}
\icmlauthor{Cong Hao}{ui}
\icmlauthor{Yuhong Li}{ui}
\icmlauthor{Yao Chen}{adsc}
\icmlauthor{Jinjun Xiong}{ibm}
\icmlauthor{Wen-mei Hwu}{ui}
\icmlauthor{Deming Chen}{ui,iot}
\end{icmlauthorlist}

\icmlaffiliation{ui}{Department of ECE, University of Illinois Urbana-Champaign, USA}
\icmlaffiliation{adsc}{Advanced Digital Sciences Center, Singapore}
\icmlaffiliation{ibm}{IBM T. J. Watson Research Center, USA}
\icmlaffiliation{iot}{Inspirit IoT, Inc., USA}

\icmlcorrespondingauthor{Xiaofan Zhang}{xiaofan3@illinois.edu}

\icmlkeywords{Machine Learning, ICML}

\vskip 0.15in
]



\printAffiliationsAndNotice{}  

\begin{abstract}
\vspace{-2pt}

Developing deep learning models for resource-constrained Internet-of-Things (IoT) devices is challenging, as it is difficult to achieve both good quality of results (QoR), such as DNN model inference accuracy, and quality of service (QoS), such as inference latency, throughput, and power consumption. Existing approaches typically separate the DNN model development step from its deployment on IoT devices, resulting in suboptimal solutions. In this paper, we first introduce a few interesting but counterintuitive observations about such a separate design approach, and empirically show why it may lead to suboptimal designs. Motivated by these observations, we then propose a novel and practical bi-directional co-design approach: a bottom-up DNN model design strategy together with a top-down flow for DNN accelerator design. It enables a joint optimization of both DNN models and their deployment configurations on IoT devices as represented as FPGAs. We demonstrate the effectiveness of the proposed co-design approach on a real-life object detection application using Pynq-Z1 embedded FPGA. Our method obtains the state-of-the-art results on both QoR with high accuracy (IoU) and QoS with high throughput (FPS) and high energy efficiency.


\end{abstract}

\vspace{-25pt}
\section{Introduction}
\vspace{-4pt}
To enable deep learning capability on IoT devices, there are two major components to be designed: the software, e.g., DNN models featuring parameters through learning for specific applications, and the hardware, such as DNN accelerators running on GPUs, FPGAs, or ASICs. Both of them contribute to the overall QoR and QoS without clear distinctions, so there is an urgent need of DNN and accelerator co-design.

\vspace{-4pt}
\subsection{Drawbacks of Independent Design Approaches}
\vspace{-2pt}
Typically, DNNs and their accelerators are designed and optimized separately for IoT applications in an iterative manner.
DNNs are first designed with more concentrations on QoR. Such DNNs can be excessively complicated for the targeted IoT devices, which must be compressed using quantization, network pruning, or sparsification \cite{wang2018design,han2017ese} before implementing on hardware, and then be retrained to maintain inference accuracy. Since no hardware constraints are captured during DNN design, this design methodology can only expect hardware accelerators to deliver good QoS through later optimizations on hardware prospects. 
%
On the other hand, the DNN accelerator design usually uses a consistent overall architecture (such as the recurrent \cite{aydonat2017opencl,zeng2018framework,jouppi2017datacenter} or pipelined structure \cite{li2016high,zhang2018dnnbuilder}) but various scale-down factors to meet different hardware constraints. 
%
%
When facing strict hardware constraints, scaling-down the accelerator is not always feasible as the shrinking resources can significantly slow down the DNN inference process and result in poor QoS. 
Design opportunities must turn to the algorithm side and ask for more compact DNN models.

\vspace{-6pt}
\section{Empirical observations}
\vspace{-4pt}
\subsection{Challenging HW/SW Configurations}
\vspace{-2pt}
One of the most fundamental barriers blocking the DNN and accelerator design is the different sensitivities of DNN/accelerator configurations (e.g., DNN model size, hardware utilization features). 
It is hard to balance these configurations using separated DNN/accelerator design approach, since a negligible change in DNN models may cause huge differences in hardware accelerators and vice versa, resulting in difficult trade-off between QoR and QoS. 

\begin{figure}[h!]
\centering
\includegraphics[width=0.53\textwidth]{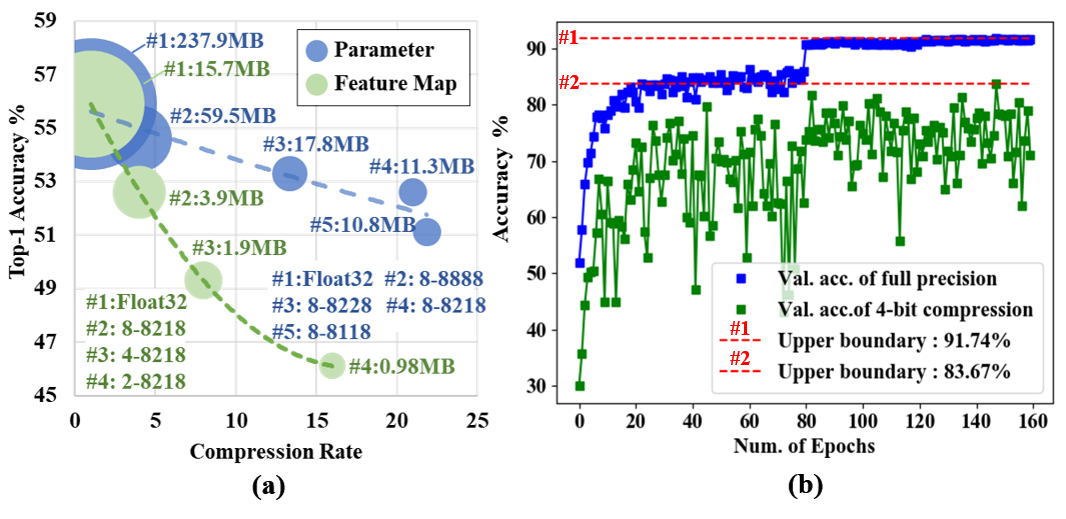}
\vspace{-25pt}
\caption{(a) Accuracy trends of AlexNet inference in ImageNet dataset during parameter (blue) and feature map (green) compression with retraining. Model name is donated as precision $p_1$ for FM, $p_2$ for the 1st CONV, $p_3$ for the 2nd$\sim$5th CONVs, $p_4$ for the 1st$\sim$2nd FCs, and $p_5$ for the 3rd FC in $p_1$-$p_2p_3p_4p_5$ format; (b) Training of ResNet-20 in Cifar10 dataset using ADMM with full-precision (blue) and quantized (green) FMs and parameters.}
\vspace{-8pt}
\label{fig:comp_train}
\end{figure}

\textit{Observation 1: similar compression rate but different accuracy.} When designing DNNs for IoT applications, it is inevitable to perform model compression. Although the overall QoS may be the same for a DNN with similar compression rates, the compression of different DNN components may cause great differences in QoR. As shown in Fig.~\ref{fig:comp_train} (a), the accuracy trends vary significantly for quantizing parameters and intermediate feature maps (FMs). In this figure, the coordinates of the bubble center represent accuracy and model compression rate, while the area shows data size in megabyte (MB). We scale-up the bubble size of FM for better graphic effect. By compressing the model from full-precision (float32) to 8-bit, 4-bit fixed point, ternary and binary representations, we reduce 22X parameter size (237.9MB$\rightarrow$10.8MB) and 16X FM size (15.7MB$\rightarrow$0.98MB), respectively. Results show that the inference accuracy is more sensitive to the precision of FM (9.8\% accuracy drop with 16X compression) compared to the parameters (4.8\% accuracy drop with 22X compression). 
Challenges also come from the difficulty of DNN training. As shown in Fig. \ref{fig:comp_train} (b), the accuracy growth of compressed model is quite unstable compared to the original full-precision model. It requires more efforts to design the training process (e.g., fine-tuning the training set-up or iteratively modifying the DNN compression rate) and more powerful machines (e.g., computer cluster for faster training \cite{PipeSGD}).

\begin{figure}[t]
\centering
\includegraphics[width=0.52\textwidth]{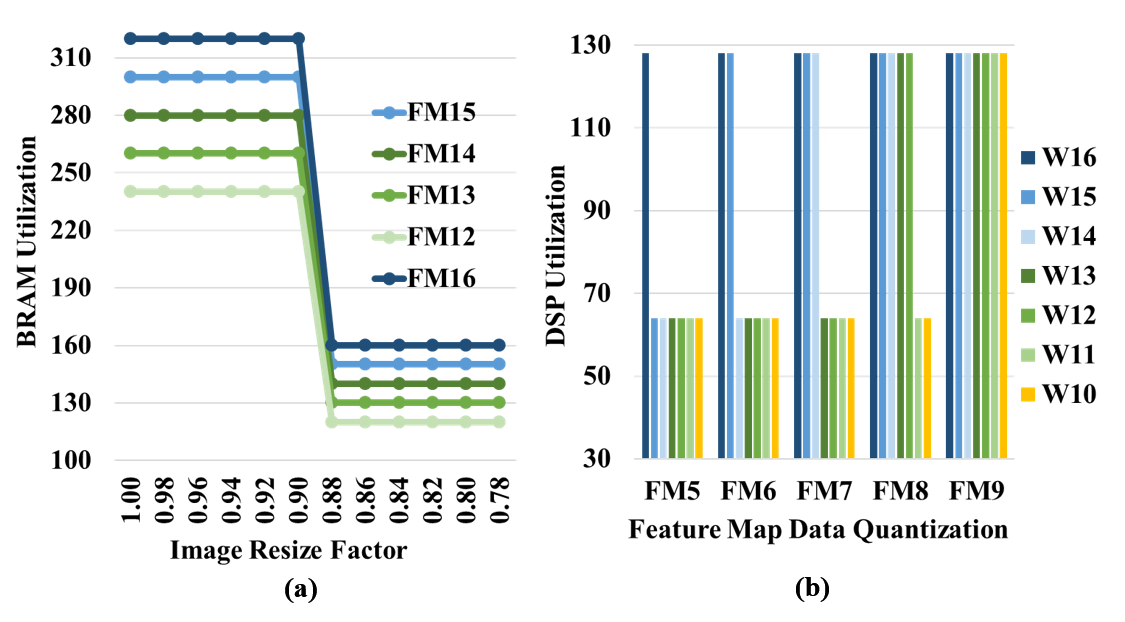}
\vspace{-26pt}
\caption{(a) BRAM usages of accelerators with the same architecture but 12$\sim$16-bit quantization for feature maps (FM12$\sim$FM16)  and different image resize factors. (b) DSP utilization of accelerator using different quantizations between weights (W) and feature maps (FMs) with the numbers indicating bits allocated.}
\vspace{-12pt}
\label{fig:hw-utilization}
\end{figure}

\textit{Observation 2: similar accuracy but different hardware resource utilization.} DNN models with similar QoR may also result in greatly different QoS because of different hardware resource usage.
%
Taking the implementation of a DNN accelerator on FPGA as an example, a single bit difference in data representation may result in considerable impacts on hardware resource utilization. 
Fig.~\ref{fig:hw-utilization} (a) shows BRAM (on-chip memory in FPGA) usage under different image resize factors with 12$\sim$16-bit data precision. By reducing the resize factor from 1.00 to 0.78, we can maintain nearly the same DNN accuracy (\textless1.0\% drop), but can save half memory when the resize factor is smaller than 0.9. 
Similarly, Fig.~\ref{fig:hw-utilization} (b) indicates that different quantization combinations of DNN feature maps and weights can result in great diverse DSP utilization.
Taking the 6-bit feature map as an example, the DSP usage reduces from 128 to 64 when weights are changed from 15-bit to 14-bit.
The reason behind is the limited bit-width support of each DSP to perform a two-input multiplication. If the bit-width of two inputs exceed a certain value, more DSPs are concatenated to handle one multiplication, which can easily double the resource utilization.
These observations imply that the configuration of hardware/software can cause great challenges of delivering desired QoR and QoS on IoT devices.

\vspace{-6pt}
\subsection{Confusing QoR Upper-Bounds for Given Tasks}
\vspace{-2pt}

When deploying DNNs on IoT devices, it is common to first find a DNN with desired QoR upper-bound for the targeted application, and then to prune the DNN to make up for the lost QoS on hardware. This solution assumes that complicated DNNs with more parameters always deliver higher QoR than simple DNNs with less parameters. However, it is not always true. By examining a UAV-based object detection task \cite{DACSDC}, we observe an abnormal trend regarding model size and QoR upper-bound (Table \ref{tab:upper-bound}), where DNNs with more parameters fail to deliver higher accuracy after adequate training. 
This implies that the current separated DNN/accelerator design may only reach suboptimal solutions, and requires more time and efforts for iterative refining before delivering prefect QoR and QoS.

\begin{table}[t]
\vspace{-10pt}
\caption{DNNs for single object detection for 3$\times$160$\times$360 input images using different backbones listed (without fully-connected layers) but the same back-end for bounding box regression. }
\label{tab:upper-bound}
\begin{center}
\begin{footnotesize}
\begin{tabular}{|l|c|c|}
\hline
Backbone & Para. Size (MB) & IoU \\
\hline
ResNet-18 \cite{he2016deep} & 85  & 61\% \\ 
ResNet-32 \cite{he2016deep} & 162  & 26\% \\ 
ResNet-50 \cite{he2016deep} & 179  & 32\% \\ 
VGG-16 \cite{simonyan2014very}  & 56  & 25\% \\ \hline

\end{tabular}
\end{footnotesize}
\end{center}
\vskip -0.3in
\end{table}

\begin{figure*}[t]
\centering
\vspace{-4pt}
\includegraphics[width=0.73\textwidth]{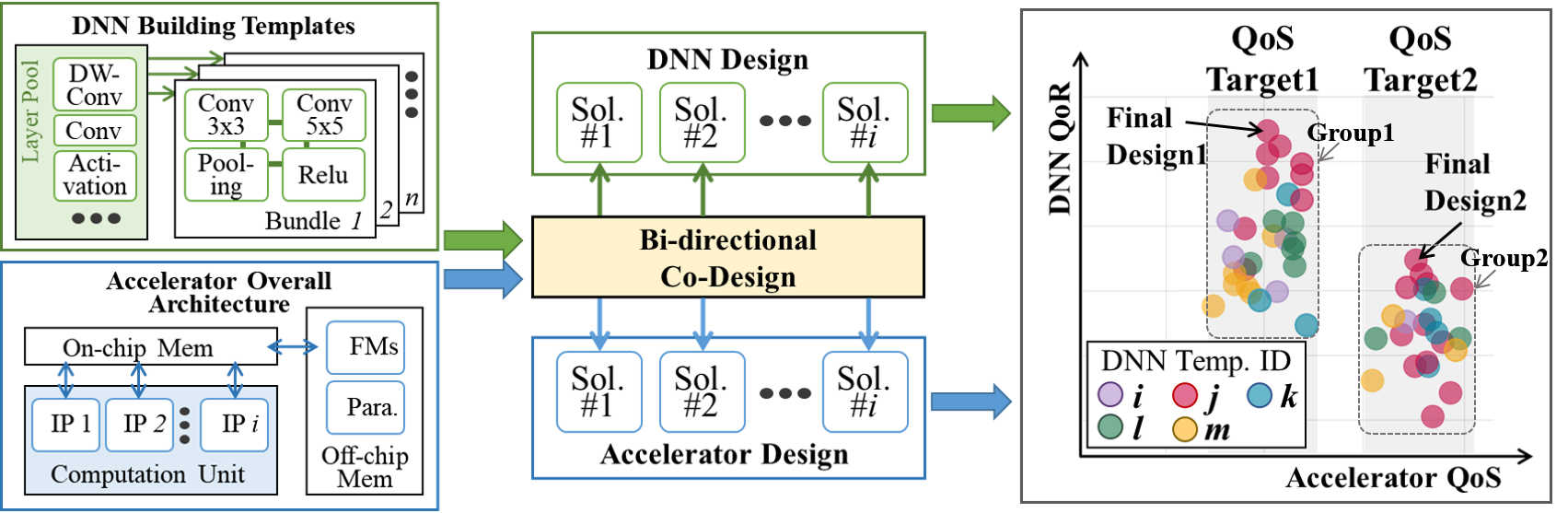}
\vspace{-8pt}
\caption{The proposed bi-directional co-design with a bottom-up DNN model exploration and a top-down accelerator design approach. For DNN exploration, we start using the hardware-aware building templates (called Bundles), and grow the DNN to reach
desired QoR; For accelerator design, we follow the proposed architecture using bundle-reused tile-based pipeline, and optimize configurable parameters to pursue the targeted QoS. }
\vspace{-12pt}
\label{fig:flow}
\end{figure*}

\vspace{-12pt}
\section{The Proposed Bi-Directional Co-Design}
\vspace{-2pt}
Motivated by the discussed observations,
we propose a bi-directional co-design methodology with a bottom-up hardware-oriented DNN design, and a top-down
accelerator design considering DNN-specific characteristics. Both DNNs and accelerators are designed simultaneously to pursue the best trade-off between QoS and QoR. The overall flow of the proposed co-design is shown in Fig. \ref{fig:flow}. The inputs of this flow include the targeted QoS, QoR, and the hardware resource constraints; the outputs include the generated DNN model and its corresponding accelerator design. We break down the whole flow into three steps:

\textit{Step1: Bundle construction and QoS evaluation.} We randomly select DNN components from the layer pool and construct bundles (as basic building blocks of generated DNNs) with different layer combinations. Each of the bundle is evaluated by analytical models to capture the hardware characteristics (e.g., latency, computation and memory demands, resource utilization), which allows QoS estimation in the early stage during DNN exploration. 

\textit{Step2:  QoR- and QoS-based bundle selection.} To select the most promising bundles, we first evaluate the QoR potential of each bundle by replicating such bundle $n$ time to construct a prototype DNN. All prototype DNNs are fast trained (20 epochs) directly on the targeted dataset for accuracy results. Based on the QoS estimation in \textit{step1}, we group prototype DNNs with similar QoS to the input targets and select the top-$n$ bundle candidates of each group.

\textit{Step3: Hardware-aware DNN exploration.} By stacking the selected bundle, we start exploring DNNs with a bottom-up approach under given QoS and QoR constraints by using stochastic coordinate descent (SCD). DNNs output from SCD are precisely evaluated regarding their QoS and fed back to SCD for DNN model update. The generated DNNs that meet QoS targets are output for training and fine-tuning to have improved QoR.

We propose a DNN accelerator which provides a tile-based pipelined architecture for efficient implementation of DNN applications with maximum resource sharing strategy. It includes a folded structure to compute DNN bundles sequentially by reusing the same hardware computing components for resource saving when targeting compact IoT devices. To ensure better QoS, it also uses an unfolded structure for computing operations (partitioned by tiles) inside bundles in a pipelined manner. With the combination of folded and unfolded structure, the proposed architecture can acquire advantages from both recurrent and pipelined structure.


\begin{table}
\vspace{-12pt}
\caption{The proposed DNNs with different data precisions for \textbf{W}eight and \textbf{F}eature map. The convolutional layers include depth-wise (DW) 3$\times$3 and point-wise (PW) 1$\times$1 convolution with output channel number shown in the bracket.}
\footnotesize   
\label{tab:propose-dnn}
\begin{center}
\begin{tabular}{|c|c|c|}
\hline
A (W16, F8)  & B (W16, F16)  & C (W11, F8)  \\ \hline
\multicolumn{3}{|c|}{input (3$\times$160$\times$360 color image)} \\ \hline
\multicolumn{3}{|c|}{\begin{tabular}[c]{@{}c@{}}DW-Conv3 (3) \\ PW-Conv1 (48)\end{tabular}} \\ \hline
\multicolumn{3}{|c|}{2$\times$2 max-pooling} \\ \hline
\multicolumn{3}{|c|}{\begin{tabular}[c]{@{}c@{}}DW-Conv3 (48) \\ PW-Conv1 (96)\end{tabular}} \\ \hline
\multicolumn{3}{|c|}{2$\times$2 max-pooling} \\ \hline
\multicolumn{3}{|c|}{\begin{tabular}[c]{@{}c@{}}DW-Conv3 (96) \\ PW-Conv1 (192)\end{tabular}} \\ \hline
\multicolumn{3}{|c|}{2$\times$2 max-pooling} \\ \hline

\begin{tabular}[c]{@{}c@{}}
DW-Conv3 (192) \\ PW-Conv1 (384) \\ PW-Conv1 (10) \\ \\ \end{tabular} & \begin{tabular}[c]{@{}c@{}}DW-Conv3 (192) \\ PW-Conv1 (384) \\ PW-Conv1 (10) \\\\ \end{tabular} & \begin{tabular}[c]{@{}c@{}}DW-Conv3 (192) \\ PW-Conv1 (384) \\DW-Conv3 (384) \\ PW-Conv1 (512)\\PW-Conv1 (10) \\
\end{tabular} \\ \hline
\multicolumn{3}{|c|}{Back-end for bounding box regression}  \\ \hline
\end{tabular}
\end{center}
\vspace{-16pt}
\end{table}

\begin{table}[h]
\vspace{-0pt}
\caption{Result comparisons to the champion design in FPGA and GPU track of DAC'18 System Design Contest \cite{DACSDC}}
\label{tab:result}
\begin{center}
\begin{small}
\begin{tabular}{|c|c|c|c|}
\hline
Model & IoU & FPS & Efficiency \\
\hline
The proposed DNN-A & 59.3\% & \textbf{29.7} & \textbf{12.38} image/watt  \\ 
The proposed DNN-B & 61.2\% & 22.7 & 9.46 image/watt \\ 
The proposed DNN-C & 68.6\% & 17.4 & 6.96 image/watt \\ \hline
Modified SSD (FPGA)  & 62.4\%  & 12.0 &  2.86 image/watt\\ \hline
Modified Yolo (GPU)  & \textbf{69.8\%}  & 24.6 &  1.95 image/watt\\ \hline

\end{tabular}
\end{small}
\end{center}
\vskip -0.2in
\end{table}

\vspace{-6pt}
\section{Results and Conclusions}
\vspace{-2pt}

We demonstrate the proposed bi-directional co-design on a real-life object detection task in DAC'18 System Design Contest and generate three DNNs (A, B, and C in Table \ref{tab:propose-dnn}) and corresponding accelerators on Pynq-Z1 FPGA for different QoS-QoR combinations. The proposed co-design flow first identifies that the bundle with DW-Conv3, PW-Conv1, and max-pooling layers is the most promising building template for the target hardware device and application. Based on this bundle, the co-design explores three DNN configurations with different quantization schemes to satisfy the QoR demands, respectively. 
As shown in Table \ref{tab:result}, we can deliver the best FPS (29.7) and efficiency (12.38 image/watt) using the same FPGA as the FPGA champion design. Among them, the proposed DNN-C outperforms the FPGA winning design in all aspects with 6.2\% higher IoU, 1.45X higher FPS, and 2.4X higher efficiency.
Comparing to the GPU winning design, the DNN-C design can deliver comparable accuracy but 3.6X higher efficiency.


\clearpage

\section*{Acknowledgment}
\quad This work was partly supported by the IBM-Illinois Center for Cognitive Computing System Research (C$^3$SR) -- a research collaboration as part of IBM AI Horizons Network.

\bibliography{example_paper}
\bibliographystyle{icml2019}


\end{document}